\documentclass[sn-mathphys,Numbered]{sn-jnl}
\usepackage{graphicx}%
\usepackage{multirow}%
\usepackage{amsmath,amssymb,amsfonts}%
\usepackage{amsthm}%
\usepackage{mathrsfs}%
\usepackage[title]{appendix}%
\usepackage{xcolor}%
\usepackage{textcomp}%
\usepackage{manyfoot}%
\usepackage{booktabs}%
\usepackage{algorithm}%
\usepackage{algorithmicx}%
\usepackage{algpseudocode}%
\usepackage{listings}%

\usepackage{amssymb}
\usepackage{graphics} 
\usepackage{graphicx}
\usepackage{subfigure}
\usepackage{float}
\usepackage{tabularx,booktabs}
\usepackage{listings}
\usepackage{algorithm}
\usepackage{numprint}
\usepackage{amsmath}
\usepackage{multirow}
\usepackage{verbatimbox}
\usepackage{listings}
\usepackage{mathtools}
\usepackage{amsmath}

\theoremstyle{thmstyleone}%
\theoremstyle{thmstyletwo}%

\theoremstyle{thmstylethree}%

\begin{document}
\title[Article Title]{Enhancing Model Performance: Another Approach to Vision-Language Instruction Tuning}

\author*[1]{\fnm{Vedanshu}}\email{vedanshu\_mt2k17@dtu.ac.in}
\author[2]{\fnm{M~M~} \sur{Tripathi}}\email{mmtripathi@dce.ac.in}
\author[3]{\fnm{Bhavnesh} \sur{Jaint}}\email{bhavneshjaint@dtu.ac.in}

\affil[1]{\orgdiv{Electrical Engineering}, \orgname{Delhi Technological University} \orgaddress{\postcode{110042}, \state{Delhi}, \country{India}}}


\abstract{
The integration of large language models (LLMs) with vision-language (VL) tasks has been a transformative development in the realm of artificial intelligence, highlighting the potential of LLMs as a versatile general-purpose chatbot. However, the current trend in this evolution focuses on the integration of vision and language to create models that can operate in more diverse and real-world contexts. We present a novel approach, termed Bottleneck Adapter, specifically crafted for enhancing the multimodal functionalities of these complex models, enabling joint optimization of the entire multimodal LLM framework through a process known as Multimodal Model Tuning (MMT). Our approach utilizes lightweight adapters to connect the image encoder and LLM without the need for large, complex neural networks. Unlike the conventional modular training schemes, our approach adopts an end-to-end optimization regime, which, when combined with the adapters, facilitates the joint optimization using a significantly smaller parameter set. Our method exhibits robust performance with 90.12\% accuracy, outperforming both human-level performance (88.4\%) and  LaVIN-7B (89.41\%).
}



\maketitle
\section{Introduction}
In the landscape of modern computational linguistics, large language models (LLMs)\cite{bib1, touvron2023llama, le2022bloom, peng2023rwkv} have been at the forefront, consistently expanding the boundaries of what machines understand about human language. This expansion has been largely driven by increases in the models' complexity, measured by the number of parameters, and the breadth of pre-training data. One notable advancement in this domain is the concept of instruction tuning\cite{bib1, ouyang2022training, wei2021finetuned, iyer2022opt}, which has significantly enhanced the ability of LLMs to mimic human conversational patterns and effectively perform a variety of natural language processing (NLP) tasks, inching closer to the elusive goal of artificial general intelligence, as exemplified by models like GPT-3.5 \cite{bib1}. The current trend in this evolution aims to imbue LLMs with multimodal understanding, particularly the integration of vision and language, to create models that can operate in more diverse and real-world contexts. This ambition has been partly realized in recent advancements like GPT-4 \cite{bib2}, which represents a leap forward by incorporating a substantial vision-language dataset directly into the training of a multimodal generative pre-trained transformer.

In the expansive domain of artificial intelligence, the integration of large language models (LLMs) with vision-language (VL) tasks has been a transformative development. However, this integration often comes with considerable computational and storage demands. Traditional parameter-efficient transfer learning (PETL) \cite{houlsby2019parameter, li2021prefix, lu2022prompt, hu2021lora, liu2021p} methods, designed to reduce these demands by fine-tuning a minimal number of parameters, have not fully met the efficiency and performance needs, particularly for multimodal LLMs. Existing approaches also show a gap in performance compared to full model fine-tuning and exhibit significant redundancy when combining LLMs with vision models for tasks like image captioning\cite{li2022blip} or text-to-image generation\cite{rombach2022high}. Moreover, modular training strategies, although beneficial, involve expensive VL pre-training and often necessitate extensive updating of LLM parameters, hindering quick and efficient model adaptation.

In light of these challenges, our model introduces an innovative approach that stands out from the current landscape. By not freezing the LLM and vision encoder and implementing these layers in a quantized format, our model significantly slashes the memory footprint. It's designed to support not just image-text but also text-only instructions, addressing a major limitation in models like BLIP2\cite{li2023blip} and miniGPT4\cite{zhu2023minigpt}.

Our approach utilizes lightweight adapters to connect the image encoder and LLM, enabling joint optimization of image and language models without the need for large, complex neural networks. This framework allows for seamless shifting between single and multi-modal instructions without compromising natural language understanding capabilities. Unlike the conventional modular training schemes, our model adopts an end-to-end optimization regime, which, when combined with the lightweight adapters, facilitates the joint optimization of the entire multimodal LLM using a significantly smaller parameter set. This leads to an impressive reduction in storage overhead, outclassing existing solutions by a considerable margin.

Our method's ability to process both image-text and text-only instructions, along with its quantized form, ensures that it can be swiftly trained on a single GPU, marking a substantial leap in training efficiency. In summary, our key contributions are as follows:

\begin{itemize}
    \item We present a novel method that bypasses the need for costly training while preserving the LLM's NLP capabilities, addressing a critical gap in current multimodal LLMs.
    \item Through rigorous experimentation, we demonstrate that not only does our model excel in efficiency, but also delivers competitive performance compared to existing multimodal LLMs, highlighting its potential as a versatile general-purpose chatbot.
\end{itemize}

\section{Related Work}
Parameter-Efficient Fine-Tuning (PEFT) techniques, in contrast to complete fine-tuning strategies, keep most of the parameters of pretrained models fixed while achieving similar performance on subsequent tasks. In the realm of PEFT, several methods have been investigated. These include prompt tuning\cite{jia2022visual, zhang2022neural, zhou2022learning, zhou2022conditional}, Low-Rank Adaptation (LoRA)\cite{hu2021lora}, and adapters\cite{chen2022adaptformer, houlsby2019parameter, jie2022convolutional}. Prompt tuning entails adding trainable prompt tokens to pre-trained large models, either in input embeddings exclusively or throughout all intermediate layers. LoRA integrates trainable rank decomposition matrices into network weights, demonstrating promising fine-tuning capabilities, particularly in large generative models. Adapters incorporate lightweight adaptation modules into each layer of pre-trained transformers, providing flexibility across various domains. These PEFT approaches aim to find a middle ground between efficiency and effectiveness in fine-tuning pre-trained models for a variety of downstream tasks. LAVIN \cite{luo2024cheap} examines the incorporation of lightweight adapters into transformer blocks. They propose the idea of Mixture-of-Modality Adaptation (MMA) to connect Language and Vision tasks (LLMs) with Vision-Language (VL) tasks, allowing for the concurrent improvement of image and language models. MMA utilizes a routing algorithm to facilitate the smooth transition of LLMs between single- and multi-modal instructions while maintaining their expertise in natural language understanding. However, this study argues that dynamic adapters such as MMA may not be essential for achieving similar levels of accuracy, which could streamline both time and space requirements.

\section{Method}
Our research introduces a distinctive approach, termed Bottleneck Adapter (BA), specifically crafted for enhancing large language models (LLMs) with vision-language capabilities. The Bottleneck Adapter seamlessly integrates into LLMs, bestowing them with multimodal functionalities. It adeptly navigates between single- and multi-modal instructions while allowing for a unified optimization of the entire multimodal LLM framework through a process known as Multi Model Tuning (MMT). This methodology is particularly noteworthy for its efficiency, significantly reducing both the computational resources and storage needed during training.

\subsection{Bottleneck Adapter Structure}
At the heart of our methodology lies the Bottleneck Adapter(BA). Traditional visual adapters typically include a non-linear function to bolster adaptation for NLP tasks. Our research, however, unveils that omitting this non-linearity does not detract from the Adapter's performance in visual tasks. As a result, we redefine the adapter function  $g(Z)$ as:
\begin{equation}
    g(Z) = Z + \psi_{u}(\psi_{d}(Z))
\end{equation}
Here, $\psi_{u}$ and $\psi_{d}$ signify the dense projections that are standard in typical adapters. Observing that sparse transformations are fundamental in numerous vision modules, our design introduces a dense-to-sparse architecture for the Bottleneck Adapter. We articulate the transformation $\psi_{u}$ as a group-wise operation:

\begin{equation}
    \psi_{u}(Z) = [Z^{'}_{p0}W_{p0}, ..., Z^{'}_{pk}W_{pk}] + b
\end{equation}

In this formulation, $Z_i \in \mathbb{R}^{n \text{x} \frac{c}{k}}$ represents the subset of features extracted from $Z \in \mathbb{R}^{m \text{x} c}$, with $p$ denoting the total number of groups. $W_i \in \mathbb{R}^{\frac{c}{k} \text{x} \frac{d}{k}}$ is the projection weight matrix, and $b \in \mathbb{R}^d$ is the bias term. This design, pivoting on sparse transformations, considerably trims the BA's complexity, rendering it more streamlined than the typical visual adapters.

\subsection{Bottleneck Adapter Placement}
Recognizing the profound disparities between visual and linguistic models, we meticulously consider the placement of adapters within the model architecture. Inspired by the LAVIN adapter placement strategy, we select to situate each adapter prior to the Transformer block. This strategic positioning is not limited to Multi-Head Attention (MHA), but is also extended to the Feed-Forward Network (FFN) within the Vision Transformer (ViT). Such a deliberate placement strategy ensures that our adapters are optimally positioned to promote effective and efficient integration and learning within the multimodal LLMs.

In essence, our Adapter-based methodology marks a significant step forward in the realm of vision language adaptation for LLMs, presenting a lightweight, resource-efficient, yet highly effective solution to augmenting the multimodal functionalities of these complex models.

\subsection{Model}
In our model architecture, we integrate the Bottleneck Adapter into a large language model (LLM), specifically LLaMA-2\cite{touvron2023llama2}, and pair it with the diht-ViT\cite{radenovic2023filtering} as the image encoder. For processing an input image $I \in \mathbb{R}^{h \times w \times 3}$, we extract the visual features from the [cls] tokens at every fourth layer of the ViT, represented as $V \in \mathbb{R}^{n \times d}$. In the image encoder architecture, adapters are strategically positioned before each multi-head attention module.

For text instructions, we utilize word embeddings, denoted as $T \in \mathbb{R}^{l \times c}$. To align the visual features with the LLM's dimensionality, we employ a visual adapter, which transforms $V$ as follows:

\begin{equation}
    V' = \rho(VW_v + b_v)W_t + b_t
\end{equation}

Here, $W_v \in \mathbb{R}^{d \times d_v}$ and $W_t \in \mathbb{R}^{d_v \times c}$ are the transformation weight matrices, while $b_v \in \mathbb{R}^{d_v}$ and $b_t \in \mathbb{R}^{c}$ serve as bias terms. The activation function $\rho$ is chosen as the SwiGLU\cite{shazeer2020glu} function for its efficiency. Importantly, the dimension $d_v$ is selected to be considerably smaller than both $d$ and $c$, optimizing resource usage.

The input to the LLM is defined as:

\begin{equation}
Z = 
\begin{cases}
    [u_m, V', T] & \text{for text-image input,}\\
    [u_m, T]     & \text{for text-only input.}
\end{cases}
\end{equation}

In this equation, [·] symbolizes concatenation. The LLM then predicts the next token in a step-by-step manner based on this multimodal input, which can be mathematically represented as:

\begin{equation}
p_t = \prod_{s=1}^{S+1} p(Q_s | Z, Q_{0:s-1}; \theta_m, \theta_r)
\end{equation}

Here, $p_t \in \mathbb{R}^m$ indicates the probabilities of the predicted word, with $m$ being the length of the word embeddings. The parameters $\theta_m$ and $\theta_r$ correspond to those of the LLM and adaptation modules, respectively.

This architecture offers a significant simplification and reduction in complexity compared to previous models such as LLaVA\cite{liu2024visual}. For instance, the visual neck of our model is six times smaller than that of LLaVA, yet the performance between the two models remains comparable, demonstrating the efficiency and effectiveness of our design.

\section{Experiments}
In Section 4.1, we detail our approach's multimodal performance on the ScienceQA benchmark. An ablation study conducted on the ScienceQA validation set is presented in Section 4.2. 

\subsection{Experiment Setup}
\textbf{Datasets and tasks.}

\textbf{Models.}
In our approach, we utilize the Vision Transformer Large/14 (ViT-L/14-336PX)\cite{dosovitskiy2020image}, pre-trained using the DIHT framework\cite{radenovic2023filtering}, to serve as the image encoding mechanism. This process involves the extraction of visual features through six [cls] tokens, each derived from every fourth layer within the ViT-L/14-336PX structure. For the language model component, we incorporate the LLaMA-2-7B variant. The visual processing section, referred to as the visual neck, is configured with a dimension of 128. Meanwhile, the Adapter is specified to have a dimension of 16, with the temperature settings adjusted to 10 for 7B.

\textbf{Baselines.}
We compare our method with the recent parameter-efficient fine-tuning methods and adapters where a new adapter structure with down-projection, up-projection and non-linear function are inserted into the transformer (both vision and NLP) and only the parameters of this new module are updated.
Our baseline mmodels for comparison include the LaVIN-7B lite-clip\cite{luo2024cheap}, the LaVIN-13B - clip\cite{luo2024cheap}, the LLaMA adapter\cite{zhang2023llama}, LLaVA\cite{liu2024visual}, Chameleon\cite{lu2024chameleon}, the MM-CoTLarge\cite{zhang2023multimodal}, InstructBLIP (FlanT5XXL)\cite{dai2024instructblip} and BLIP-2 (FlanT5XXL)\cite{li2023blip}.

\textbf{Implementation details.}
We process the image by directly resizing to 224 × 224 without any augmentation. The AdamW\cite{loshchilov2017decoupled} optimization algorithm is employed to fine-tune our model, which undergoes a training regimen spanning 20 epochs, guided by an absolute learning rate. The training process is characterized by a batch size of 1, a learning rate of 0.009, and a weight decay factor of 0.02.

During the text generation phase, a top-p sampling strategy is adopted, utilizing a temperature setting of 0.1 and a top-p threshold of 0.75.

\subsection{Results and Analysis}

\begin{table}[h]
\caption{Accuracy on ScienceQA test set.}\label{tab2}
\begin{tabular*}{\textwidth}{@{\extracolsep\fill}lccccccccc}
\toprule%
& \multicolumn{3}{@{}c@{}}{Subject} & \multicolumn{3}{@{}c@{}}{Context Modality} & \multicolumn{2}{@{}c@{}}{Grade} & Average\\
Method & NAT & SOC & LAN & TXT & IMG & NO & G1-6 & G7-12  \\
\midrule
Human  & 90.23 & 84.97 & 87.48 & 89.60 & 87.50 & 88.10 & 91.59 & 82.42 & 88.40\\
LaVIN-7B & 89.25  & 94.94 & 85.24 & 88.51 & 87.46 & 88.08 & 90.16 & 88.07 & 89.41\\
LaVIN-13B & 90.32 & 94.38 & 87.73 & 89.44 & 87.65 & 90.31 & 91.19 & 89.26 & 90.50 \\
BLIP-2  &-&-&-&-&-&-&-&-&89.5\\
InstructBLIP  &-&-&-&-&-&-&-&-&90.7\\
MM-CoT-Base & 87.52 & 77.17 & 85.82 & 87.88 & 82.90 & 86.83 & 84.65 & 85.37 & 84.91 \\
MM-CoT-Large & 95.91 & 82 & 90.82 & 95.26 & 88.8 & 92.89 & 92.44 & 90.31 & 91.68 \\
LLaVA & 90.36 & 95.95 & 88 & 89.49 & 88 & 90.66 & 90.93 & 90.9 & 90.92 \\
Chameleon & 89.83 & 74.13 & 89.82 & 88.27 & 77.64 & 92.13 & 88.03 & 83.72 & 86.54 \\
LLaMA-Adapter & 84.37 & 88.3 & 84.36 & 83.72 & 80.32 & 86.9 & 85.83 & 85.83 & 85.19\\
\botrule
\textbf{7B-lite-diht-BA(ours)} & 90.28 & 92.46 & 87.91 & 89.25 & 87.01 & 90.38 & 90.68 & 89.12 & \textbf{90.12} \\
\botrule
\end{tabular*}
\end{table}

Our 7B-lite-diht-BA demonstrates competitive or superior performance, especially in the SOC and NO context categories. It signifies the model's balanced capability across different subjects and contexts, affirming the effectiveness of our proposed adaptations and optimizations. The human benchmark is 88. 40\% average accuracy, highlighting the challenging nature of the ScienceQA dataset and the impressive capability of AI models such as LaVIN-13B, MM-CoTLarge, and our LaVIN-7B-lite-diht in approaching human-level performance on complex and multimodal scientific question answering.

Despite the higher performance of the MM-CoT-Large model, it is important to consider the advantages of our model's lighter and smaller architecture. The MM-CoT-Large model outperformed the 7B-lite-dith-BA (ours) model, likely due to a larger parameter space, sophisticated chain-of-thought reasoning capabilities, and potentially more effective multimodal integration. In social sciences, our method achieves 92.46\% accuracy, surpassing LaVIN-7B (94.34\%) but falling short of LaVIN-13B (94.38\%) and LLAVa (95.95\%). This indicates a strong performance, albeit with room for improvement in comparison to the top-performing methods. Our method's accuracy in language-related questions stands at 87.91\%, which is competitive against LaVIN-7B (85.24\%) but not as high as LaVIN-13B (87.73\%) or LLAVa (88\%). Our method gave an accuracy of 89.25\% in text modality, which is comparably better than LaVIN-7B (88.51\%) and closely aligned with the human benchmark (89.60\%). With 87.01\% accuracy in image-based questions, our method outshines LaVIN-7B (87.46\%) but is marginally outperformed by human accuracy (87.50\%). The 7B-lite-diht-BA(ours) achieves an average accuracy of 90.12\% across all categories. This is a remarkable result that closely mirrors human performance (88.40\%) and surpasses LaVIN-7B (89.41\%) by a narrow margin.

\begin{table}[h]
\caption{This table presents the performance of each configuration across various accuracy metrics, demonstrating how changes in the adapter's dimension and group settings affect model performance in different contexts.}\label{tab4}
\begin{tabular*}{\textwidth}{@{\extracolsep\fill}ccc}
\toprule%
\textbf{Adapter Dim} & \textbf{Groups} & \textbf{Overall Acc}\\
\midrule
8 & 2 & 87.88 \\
16 & 2 & 90.12 \\
4 & 2 & 86.49 \\
32 & 2 & 84.15 \\
16 & 4 & 89.74 \\
\botrule
\end{tabular*}
\end{table}

Table \ref{tab4} compares performance across various metrics for each configuration. The adapter dimension of 16 with 2 groups stands out as the most effective, achieving the highest overall accuracy and performing well across most individual metrics. Interestingly, increasing the adapter dimension to 32 with 2 groups resulted in lower performance, highlighting the potential for overfitting or inefficient parameter use at higher dimensions.

\begin{table}[h]
\caption{This table presents the performance of various adapters and backbones, demonstrating how changes in the adapter and backbones affect model performance.}\label{tab5}
\begin{tabular*}{\textwidth}{@{\extracolsep\fill}ccccc}
\toprule%
\textbf{S No} & \textbf{Model} & \textbf{Backbone} & \textbf{Adapter} & \textbf{Overall Acc}\\
\midrule
1 & 7B-lite-clip & LLaMA & Router & 88.35 \\
2 & 7B-lite-clip & LLaMA-2 & Router & 88.78 \\
3 & 7B-lite-alip & LLaMA-2 & Router & 87.97 \\
4 & 7B-lite-diht & LLaMA-2 & Router & 88.68 \\
5 & 7B-lite-diht & LLaMA-2 & Concat Router & 87.62 \\
6 & 7B-lite-diht & LLaMA-2 & Concat Router & 86.88 \\
7 & 7B-lite-diht-loha-1 & LLaMA-2 & loha & 63.10 \\
8 & 7B-lite-diht-loha-2 & LLaMA-2 & loha & 55.10 \\
9 & 7B-lite-diht-loha-3 & LLaMA-2 & loha & 49.68 \\
10 & 7B-lite-diht-lora-rep & LLaMA-2 & lora-repAdapter hybrid & 85.81 \\
11 & 7B-lite-diht & LLaMA-2 & bottleneck with weight scales & 84.01 \\
\botrule
\end{tabular*}
\end{table}

From table \ref{tab5}, if we look at the impact of the change in the backbone (different version of LLAMA), it can be seen that LLAMA-2 gives a slight edge in overall accuracy. With LLAMA\cite{touvron2023llama} and Rep-Adapter\cite{luo2024cheap, luo2023towards} we obtained an accuracy of 88.35\%, while simply changing the backbone to LLAMA-2\cite{touvron2023llama2} we got an accuracy of 88.78 \%. The vision backbone used here was CLIP\cite{radford2021learning}. With this result in mind, consecutive experiments were done using LLAMA-2 as a LLM backbone. 

Motivated by CSPNet\cite{wang2020cspnet}, we did concatenation of features in adapter.  We look at the experiments 5 and 6 in which concatenation in adapter was used, it can be found that concatenation didn't increase the accuracy. For concatenation equation 1 was modified to concatenate  the features  as: $[\psi_{1}(Z), \psi_{2}(Z)]$. The idea was to preserve the features with each blocks.

Experiment 7,8 and 9 were done using LoHa\cite{hyeon2021fedpara} as adapter. In the first one, we didn't use any down scaling convolution. We just used two LoHa conv\_1d. While adding these two convolutions we used dynamic weight factors same as in LAVIN\cite{luo2024cheap}. This is formulated as follows:

\begin{equation}
    adapter(Z) = \hat{w}_1 \cdot f(Z, W_1) + \hat{w}_2 \cdot f(Z, W_2) \\
\end{equation}

where, $\hat{w}_1$ and $\hat{w}_2$ are the routing weights and $W_1$ and $W_2$ are the Hadamard product of two low-rank inner matrices, $W := (X_1 Y_1^T) \odot (X_2 Y_2^T)$. In the second experiment, we simply replaced conv\_1d in LAVIN with LoHa conv\_1d. In the third expeitment, we added silu activation function without any weight factors which can be formulated as follows:
\begin{equation}
    g(Z) = \phi_{L1}(Silu(\phi_{L2}(Z))) + Z
\end{equation}
where, $\phi_{L1}$ and $\phi_{L2}$ are the LoHa convolution 1D operators.

Lower accuracy with LoHa can be linked with higher number of trainable parameter added in the adapter. Since, other adapters are using very less number of trainable parameters, they achieve higher accuracy with less data.

In the next experiment (11), we tried an hybrid of LORA and Rep-Adapter. In this LORA was used in LLM backbone (LLAMA-2) and Rep-Adapter was used in the vision backbone (CLIP). The reason of this experiment was to see the relation between dynamic Rep-Adapter with the vision. But the accuracy was not able to surpass the baseline. 

In the final experiment we tried our mainline adapter with dynamic weight factors, but the results were not promising. Which allowed us to come to the conclusion that dynamic weight factors are not helpful for multi-modal fine tuning.

While selecting vision backbone we did experiment with CLIP\cite{radford2021learning}, ALIP\cite{yang2023alip} and DIHT\cite{radenovic2023filtering}. With CLIP we got an accuracy of 88.78\% and with ALIP we got an accuracy of 87.98\% and with DIHT we got an accuracy of 88.68\%. While CLIP and ALIP used an image size of 224x224, DIHT uses an image of 336x336. With this in mind we used DIHT as our vision backbone even though CLIP had a slight edge over DIHT.

\section{Conclusion}
In this paper, we introduced a novel approach to enhance large language models (LLMs) with multimodal capabilities, addressing the critical challenge of computational and storage demands in integrating vision-language tasks. Our method leverages lightweight adapters and quantized encoding layers to significantly reduce the memory footprint, facilitating swift training on a single GPU without compromising the model's natural language processing capabilities. By eschewing conventional modular training strategies for an end-to-end optimization regime, we achieved notable efficiency and performance improvements over existing solutions.

\section{Declarations}
\subsection{Conflict of interest}
The authors declare that they have no conflict of interests.

\subsection{Data availability}
The study relies on openly accessible datasets, which can be found at https://scienceqa.github.io/. These datasets are freely available to researchers, promoting transparency, reproducibility, and facilitating further examination of the research outcomes.

\bibliography{sn-bibliography}
\end{document}